\documentclass[letterpaper, 10 pt, conference]{ieeeconf}  
\IEEEoverridecommandlockouts                               
\usepackage{graphicx}                       
\usepackage{graphics}                       
\usepackage{epsfig}                         
\usepackage[tight,footnotesize]{subfigure}  

\usepackage{amssymb,amsmath}
\usepackage{mdwmath}
\usepackage{commath}   
\usepackage{eqparbox}
\usepackage{mathtools}
\usepackage[utf8]{inputenc} 
\usepackage[english]{babel}
\usepackage{amsmath,amsfonts,amssymb}

\DeclareMathOperator*{\Argmax}{argmax}

\newcommand{\ie}{{\em i.e.\ }}

\newcommand{\beq}{\begin{equation}}
\newcommand{\eeq}{\end{equation}}
\newcommand{\bear}{\begin{eqnarray}}
\newcommand{\bears}{\begin{eqnarray*}}
\newcommand{\eear}{\end{eqnarray}}
\newcommand{\eears}{\end{eqnarray*}}
\newcommand{\bdm}{\begin{displaymath}}
\newcommand{\edm}{\end{displaymath}}
\newcommand{\lba}{\left[\begin{array}}
\newcommand{\ear}{\end{array}\right]}

\usepackage{enumerate}
\let\labelindent\relax
\usepackage[inline]{enumitem}

\usepackage[amsmath, thmmarks]{ntheorem} 
\theoremstyle{plain}
\newtheorem{theorem}{Theorem}
\newtheorem{corollary}[theorem]{Corollary}
\newtheorem*{comment}{Comment}

\usepackage{stfloats}                       
\usepackage{hyperref}
\usepackage{cite}                           
\usepackage[T1]{fontenc} 
\usepackage{tabularx}
\usepackage{multirow}
\usepackage[table]{xcolor}
\usepackage{longtable}
\usepackage{booktabs} 			
\usepackage{xcolor,colortbl}    
\newcommand{\red}{\textcolor[rgb]{ .773,  0,  .043}}

\newcommand\given[1][]{\:#1\vert\:}

  \title{\LARGE \bf Fast, Robust, and Versatile Event Detection through HMM Belief State Gradient Measures.}
\author{Shuangqi Luo, Hongmin Wu, Hongbin Lin, Shuangda Duan, Yisheng Guan, and Juan Rojas. \\
}
\begin{document}
\maketitle
\thispagestyle{empty}
\pagestyle{empty}
\bstctlcite{IEEEexample:BSTcontrol}
\begin{abstract}
Event detection is a critical feature in data-driven systems as it assists with the identification of nominal and anomalous behavior. Event detection is increasingly relevant in robotics as robots operate with greater autonomy in increasingly unstructured environments. In this work, we present an accurate, robust, fast, and versatile measure for skill and anomaly identification. A theoretical proof establishes the link between the derivative of the log-likelihood of the HMM filtered belief state and the latest emission probabilities. The key insight is the inverse relationship in which gradient analysis is used for skill and anomaly identification. Our measure showed better performance across all metrics than all but one related state-of-the-art works. The result is broadly applicable to domains that use HMMs for event detection. 
Supplemental information, code, data, and videos can be found at \cite{2018ROMAN-Luo-EventDetection_supplementalURL}.
\end{abstract}
\section{INTRODUCTION}\label{sec:Intro}
%
%
Event detection is a key component of data-driven systems where maintaining a belief about its state outcome (nominal or anomalous) is imperative for long-term autonomy. In robotics, event detection, is increasingly critical as robots operate with greater autonomy in increasingly unstructured environments.

Event detection has traditionally focused on anomaly detection in industrial tasks such as parts assembly \cite{1982AI-Brooks-SymbolicErrorAnalysisBotPlanning}, \cite{1999Patent-Fullmer-PartsAsmblySigAnlysis}. With increasing access to data, robotics has turned to data driven methods \cite{2005BotAut-Petterson-ExecutionMonitoringInRoboticsSurvey}. With increasing computation, sensor, and actuator resources, continuous multi-modal signals are used along with better robot and environmental modeling for better event detection  \cite{2013IROS-DiLello-BayesianContFaultDetection,2015ICRA-Kroemer-TwrdsLearnHierSkillsMultiPhaseManip,2017humanoids-rojas-shdp-var-hmm,2017IROS-Park-MultiModalAnomalyClassifFeeding,2018ROMAN-Wu-RecovExtDist_StateDep,2016ISER-Levine-LearnHandEyeCalib_BotGrasping_DL}. Recently, there has been interest in not just identifying anomalies but also incrementally classifying them and integrating recovery mechanisms into their systems \cite{2015IJRR-Niekum-LrnGrnddFiniteStateReprUnstrucDems,2018ROMAN-Wu-RecovExtDist_StateDep}. The goal is to close loops between high-level event detectors and low-level robot controllers that can optimally adjust or recover from anomalous events. Not much work has been done in the study of event detection in the post-anomaly-recovery stage of a task \cite{2015RSS-Kappler-DateDrivenOnlineDecisionMakingManipu,2018ROMAN-Wu-RecovExtDist_StateDep}. Little quantitative and qualitative analysis is available for such techniques, which are critical for long-term autonomy. 

This paper studies the development of an event detection measure useful for nominal and anomaly identification, and one that is robust (even) in post-anomaly-recovery scenarios where false-positives are easily triggered. Our contribution presents a theoretical proof for an event detection metric derived from the gradient computation of the natural logarithm of the HMM filtered belief state (from hereon referred to as the ``forward gradient'' measure) as well as a key guarantee for the admissibility of the measure. The measure provided very good identification accuracy, robustness across task scenarios, and fast reaction times. Extensive quantitative and qualitative analysis of the measure is also presented. 

Our theoretical derivation establishes the link between the derivative of the forward gradient and the latest emission probabilities. The result uses a set of verifiable claims and insights built on the nature of Viterbi paths and aided by the log-sum-exp trick to establish a strong correlation between observations and the forward gradient. The key insight was the inverse relationship in which gradient analysis can be used for skill and anomaly identification. We also incorporate an automated threshold that requires no manual tuning and one that is very robust under a wide variety of scenarios including post-anomaly recovery. 

A pick-and-place task, composed of five skills (basic robot manipulation actions that act as sub-goals) and multiple types of induced anomalies (deviations from nominal sensorial experiences) were used to test the performance of the measure. Anomalies were induced in varying ways across the task, including during post-anomaly recovery stages in the task. The gradient-based metric had an overall skill identification average accuracy of 98.4\%, an overall average reaction time of 1.84\% across skills (see Sec. \ref{subsec:reaction_time}), and 0 false-positive counts in pre- and post-recovery conditions (see Sec. \ref{subsubsec:pre_rec_perf} and Sec. \ref{subsubsec:post_rec_perf}). It is notable that the same gradient-based measure can be used across the board for skill and anomaly identification and in post-anomaly-recovery scenarios showing it to be very versatile. Our results had a very strong performance compared with state-of-the-art results across the board. The measure can be broadly utilized if HMMs are used for event detection.
Supplemental information, code, data, and videos can be found at \cite{2018ROMAN-Luo-EventDetection_supplementalURL}.
\section{Literature Review} \label{sec:lit_review}
This review examines the approach and effectiveness of techniques used for anomalous and nominal event identification, anomaly characterization, and identification in post anomaly-recovery.

For anomaly detection, we see that in \cite{2005IJMTM-Cho-ToolBreakageDetection,2008IJAMT-Hhsue-PredictToolBrkg_Milling_SVM}, support vector machines (SVMs) identified tool breakage detection. In \cite{2010CASE-Rodriguez-FailureDetAsmbly_ForceSigAnalysis}, SVMs identified failure in simple pick actions. The same work was rendered probabilistic later \cite{2011IROS-Rodriguez-AbortRetry} and triggered retry actions. In \cite{2017IROS-Park-MultiModalAnomalyClassifFeeding}, an HMM with an execution-variable threshold was used to identify anomalies in manipulation tasks. Deep motor learning policies like \cite{2016IJML-Abeel-End2EndTraining} do not explicitly identify anomalies though they provide inherent robustness to many perturbations. The policies still failed when significant visual differences were included in the scene.

Beyond anomaly detection, some have worked to identify manipulation skills during execution. In \cite{2015ICRA-Kroemer-TwrdsLearnHierSkillsMultiPhaseManip}, uses a state-based autoregressive HMM to model the skills and transitions of a task. In \cite{2015RSS-Kappler-DateDrivenOnlineDecisionMakingManipu}, two independent na{\"i}ve Bayes classifiers are run to identify skills and anomalies simultaneously. In \cite{2017iros-rojas-onlinewrenchintrospection}, multimodal signals were segmented into a grammar via a heuristic. The grammar was fed into an online probabilistic multi-class SVM to identify skills and anomalies. In \cite{2017humanoids-rojas-shdp-var-hmm}, a nonparametric sticky Hierarchical Dirichlet Process Vector Autoregressive HMM was used to identify skills and anomalies in snap assemblies and pick-and-place tasks.

In \cite{2008JMES-Althoefer-AutFailClassAsmblyThreadFastn}, artificial neural networks (ANNs) are used with radial basis functions to characterize insertion failures for self-tapping threaded fastenings. In \cite{2014Humanoids-Rojas-ContextualizedEArlyFailureCharac}, a probabilistic model was used to characterize different failure types in snap assemblies. In \cite{2013IROS-DiLello-BayesianContFaultDetection}, used nonparametric Bayesian hierarchical Hidden Markov Models to learn possible failure types in an alignment task. In \cite{2017IROS-Park-MultiModalAnomalyClassifFeeding}, a multi-layer perceptron composed of a temporal and a convolutional component were used to identify 12 anomalies in robot-assisted feeding.

Only a couple of these works effect recovery techniques after an anomaly is detected. In \cite{2015RSS-Kappler-DateDrivenOnlineDecisionMakingManipu}, the online decision making system is able to recover from external perturbations like human collisions. The recovery however, is performed only once for a single task and no quantitative analysis is provided for the robustness of the identification method post-recovery. This is a critical point in assessing the robustness, accuracy, versatility, and reaction speed of the technique, as conditions can change drastically in a post recovery environment from that used in training for the original identification tasks. In \cite{2018ROMAN-Wu-RecovExtDist_StateDep}, skill identification and anomaly detection were implemented through nonparametric HMM models. A generic recovery system was implemented and event detection studied after recovery actions. Our work presents a more robust technique, and one that is specially useful in post-recovery actions. We also present a theoretical proof along with a guarantee and quantitative analysis of the accuracy, robustness, and reactivity of the measure. 
\section{Problem Formulation}\label{sec:problem_formulation}
In this section we introduce Hidden Markov Models (HMMs) and how they are used for skill and anomaly identification in robotic tasks. We present weaknesses in current approaches and the motivation to find better measures. 
\subsection{Hidden Markov Models Overview}
HMMs are a doubly stochastic and generative process used to make inference on temporal data \cite{1998IJRR-Hovland-HMM_ProcessMonitorAsmbly}. The underlying stochastic process (latent states or modes) is not directly observable and represents sub-skills or actions in manipulation tasks. Latent states are observed through another set of stochastic processes that produce the sequence of observed symbols. In robotics, such observations are usually produced by noisy sensor signals (often multimodal observations). Parametric HMMs contain a finite and fixed number of latent modes which generate observations via mode-specific emission distributions (nonparametric HMMs use Bayesian techniques to learn the number of modes \cite{2017humanoids-rojas-shdp-var-hmm}). Transition distributions control the probability of transitions across latent modes over time given an initial transition probability. HMMs assume conditionally independent observations given the generative latent state. Though Markov Jump Linear Systems can model more complex dynamics and can be integrated into the HMM \cite{2017humanoids-rojas-shdp-var-hmm}. 
\subsection{Training}
For this work, single HMMs are used to model individual robot skills. HMMs can use the Baum-Welch algorithm to infer model parameters that maximize the probability of an observation given a model (many other techniques are also available see \cite{murphy2012machine}). The notation below is used to describe the HMM based on continuous observations:\\
\begin{itemize}
\item[] $Z_t$, the latent random variable at time $t$. $Z_t\in\{1,\mbox{...}, N\}$	
\item[] $z_t$, the hidden state at time \(t\) 
\item[] $\pi_i$, the initial state distribution $P(Z_1=i)$	
\item[] $A_{ji}$, the transition probability $P(Z_{t+1}=i \mid Z_{t}=j)$		
\item[] $Y_t$, the observation random variable at time \(t\)
\item[] $y_t$, the observation at time $t$
\item[] $b_i(y_t)$, the emission probability $P(Y_t=y_t \mid Z_t=i)$
\item[] $\Pi$, the HMM model composed of $\pi_i$,$A_{ji}$, $b_i(y_t)$
\item[] ${\alpha_i}(t)$, the belief state $P( Z_t=i \mid Y_{1:t}=y_{1:t}, \Pi )$
\item[] $L_t$, observation's log-likelihood $log \mbox{ } P(Y_{1:t}=y_{1:t} \mid \Pi)$.\\
\end{itemize}
We simplify notation by omitting random variable declarations: $P(Z_{1:t}=z_{1:t} \mid Y_{1:t}=y_{1:t})$ is written as $P(z_{1:t} \mid y_{1:t})$. 
\subsection{Skill Identification Methodologies}\label{subsec:previ_skill_id}
In \cite{2013IROS-DiLello-BayesianContFaultDetection,2017humanoids-rojas-shdp-var-hmm}, HMM scoring $L$ is used for skill identification. Given $S$ trained models for $S$ robot skills, scoring yields the log-likelihood of a sequence of observations at time $t$ for a trained model $s \in S$. Scoring is defined as:
\begin{equation}
    H_{t,s}=\log \mbox{ }P(y_{1:t} \mid \Pi).
\end{equation}
After scoring the models, skill selection selects the most likely candidate: \ie given a test trial $r$, the (cumulative) log-likelihood $H_{T,s}$ is computed for test trial observations conditioned on all available trained skills' model parameters $log \mbox{ } P(y_{r_1:r_t} | \Pi)_s^S$. The skill with the highest log-likelihood is selected:
\begin{equation}
	s^*=\Argmax_{s\in S}[H_{T,s}].
\end{equation}
\subsection{Anomaly Identification Methodologies}\label{subsec:PrevAnomalyIdentMethods}
Motion skill encoding is based on the premise that similar skills yield similar sensory-motor signatures \cite{2012Humanoids-Pastor-TowardsASMs, 2016ICRA-Park-MultiModalMonitoringAnomalyDet_RobotManip,2017IROS-Park-MultiModalAnomalyClassifFeeding,2017humanoids-rojas-shdp-var-hmm,2010RAM-Calinon-LrnReprodGesturesImitation}. As such, an HMM model $\Pi_s$ is derived from training data for a robot skill $s$. Optimized models (those whose scores improve as a function of selection of covariance model for Gaussian observations and latent state complexity) elicit narrower distributions. For trials belonging to the same class, the log-likelihoods of observations of the same skill yield curves that are parallel to the expected log-likelihood $\mathbb{E}(H_T)$. From these results, an anomaly threshold is devised. Often an anomaly threshold for a given skill $F1_{s_c}$ can be set as an offset from $H_T$: $F1_{s_c}=\mu(H)-k*\sigma(H)$, where $k$ is a real-valued constant that can be multiplied by the standard deviation of the expected log-likelihood to change the threshold. An anomaly is flagged if the likelihood of a test trial $r$ crosses the lower threshold: $\mbox{if } log \mbox{ } P(y_{r_1:r_t}  \given \Pi_{correct}) < F1_{s_c} \mbox{: anomaly, else nominal}$. However, such thresholds are not robust in post-recovery actions \cite{2017humanoids-rojas-shdp-var-hmm}, where numerous false-positive are triggered. At the beginning of a skill, the standard deviation $\sigma_{\mathbb{E}(L)}$ is small. So, small test observation deviations from trained observations lead to large threshold changes that trigger the false alarm. 

A second threshold definition was designed to overcome this situation (see \cite{2018ROMAN-Wu-RecovExtDist_StateDep} for details). The new threshold computed the derivative of the difference between the log-likelihood and the original anomaly threshold: $F2_{s_c}=d\mid H-F1_{s_c} \mid / dt$. This measure is robust to false-positives in post-recovery actions \cite{2018ROMAN-Wu-RecovExtDist_StateDep}; however, when the HMM model is not properly optimized, the log-likelihood curves can diverge considerably from the expectation (not parallel). The large differences in the curves affect the gradient and lead to false-positives. This work derives a more robust measure. The measure is devised from insights into the gradient of the log-likelihood function. 
\section{Theoretical Proof for Event Detection based on the Gradient Computation of HMM Log-Likelihood Data} \label{sec:study_of_gradient}
We presents a summary of the Forward algorithm and the Viterbi algorithm before presenting the gradient-based measure theory. 
\subsection{HMM Data Log-Likelihood Computation}
Given an HMM model $\Pi$ and an incoming time series $Y_{1:t}$, the natural logarithm of the filtered belief state (see 17.4.1 \cite{murphy2012machine}) associated with the forward model for latent state $i$ can be represented according to Eqtn \ref{eq:sum_of_log_alpha}.
\begin{align}
	L_t&=\log{\sum_{i=1}^{N}{\alpha_i}(t)}=\log{\sum_{i=1}^{N}{\exp({\log{{\alpha_i}(t)}})}}.
    \label{eq:sum_of_log_alpha}
\end{align}
To compute \(L_t\), we first compute \(\log{{\alpha_i}(t)}\). According to the forward-algorithm, we have:
\begin{gather}
	{\alpha_i}(1)=\pi_i {b_i}(y_1), \nonumber \\ 
	{\alpha_i}(t+1)={b_i}(y_{t+1})\sum_{j=1}^{N}{{\alpha_j}(t)A_{ji}}.
    \label{eq:recursive_computation}
\end{gather}
From Eqtns. \ref{eq:sum_of_log_alpha} \& \ref{eq:recursive_computation}, we know that \(\log{{\alpha_i}(t)}\) can be computed recursively through \(\log{{\alpha_*}(t-1)}\). Expanding the log in Eqtn. \ref{eq:sum_of_log_alpha} we have:
\begin{equation}
\log{{\alpha_i}(t)}=
\log{{b_i}(y_{t})}+\log{\sum_{j=1}^{N}{\exp({\log{{{\alpha_j}(t-1)}+\log{A_{ji}}}})}}
\label{eq:log_alpha}
\end{equation}
\subsection{Viterbi Path in HMMs}
The Viterbi algorithm, expanded in Eqtn. \ref{eq:Viterbi_recursion}, attempts to estimate the most likely state sequence. Viterbi uses dynamic programming to estimate the underlying state sequence $\hat{z}_{1:t}$ through MAP computations given a sequence of observations $y_{1:t}$ (17.4.4 in \cite{murphy2012machine}):
\begin{align}
  \hat{z}_{1:t}&=\arg\max_{z_{1:t}} P(z_{1:t} \mid y_{1:t}) \nonumber \\
  	&=\begin{aligned}[t]
    &\arg\max_{z_{t}} (\mbox{ }{b_{z_t}}(y_{t})\\
    &\quad\arg\max_{z_{1:t-1}}(A_{z_{t-1}z_t} P(z_{1:t-1}, y_{1:t-1})) \mbox{ })
	\end{aligned} \nonumber \\
	&=\begin{aligned}[t]
  	&\arg\max_{z_{t}}({b_{z_t}}(y_{t})\\
  	&\quad\arg\max_{z_{t-1}}(A_{z_{t-1}z_t}{b_{z_{t-1}}}(y_{t-1})\\
  	&\quad\quad\ddots\\
  	&\quad\quad\quad\arg\max_{z_{1}}(A_{z_{1}z_{2}}\pi_{z_1}{b_{z_1}(y_1)})\dots))
    \label{eq:Viterbi_recursion}
  	\end{aligned}
\end{align}
\begin{figure}[t]  
    \centering        
        \includegraphics[width=\linewidth]{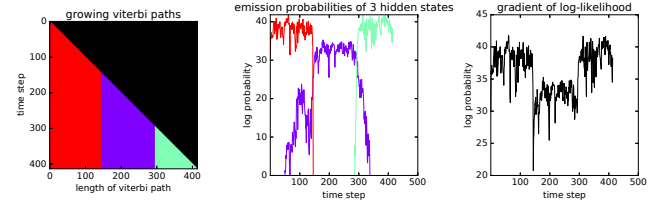}
        \caption{This figure presents 3 plot types generated by one HMM model with three hidden states. Each hidden state is represented by red, purple, and green respectively. The first plot shows the results from the Viterbi algorithm. Each row represents a time-step in the trial. With each succeeding row, a Viterbi path corresponding to the y-axis time grows. Notice, the Viterbi paths grows in a time-consistent way; that is, each Viterbi path expands on the previous one. Black pixels represent no data since the path has not grown that long. The second plot shows emission probabilities \(P(Y_t=y_t \mid Z_t=i)\) for the 3 hidden states over time \(t\). The third plot is the gradient of log-likelihood of the data computed over time \(t\). The black curves in the third plot are the maxima of the second plot, which intuitively suggests that the gradient is related to the emission probabilities.}
        \label{fig:Viterbi_triangle_emissionprob_gradient} 
\end{figure}
\subsection{The Hidden Markov Model Forward Gradient}
First, we introduce simplified notation for the Viterbi algorithm, where a node $j$ has a maximal belief state $\delta_{t}(j)=\max_{z_{1:t}} P(z_{1:t}=j \mid y_{1:t})$ with associated traceback $\hat{z_t}$.
\begin{theorem}
\label{Viterbi_path_presupposition}
For an incremental time series $Y$, a good HMM model outputs an incremental Viterbi path that \textbf{\textit{stably expands}} on the previous one. The stable expansion of the Viterbi path is as follows: given a Viterbi path "11223" for an input $Y[1\mbox{:}t]$, then the path at $Y[1\mbox{:}t+1]$ becomes "11223*", where * is the newly appended hidden state.
\end{theorem}
\begin{comment}
Good models are those that predict their data as accurately as possible and can be achieve through two steps: (i) HMM parameters optimization: the Baum-Welch BW) algorithm given a proper initialization (17.5.2 in \cite{murphy2012machine}) or similar algorithm incrementally optimize HMM parameters until a local maximum of likelihood is reached; minimizing the perplexity of the model. (ii) Model selection optimization: this consists in selecting the number of hidden states and values for observation models. Many techniques exist including BIC, MCMC, Variational Bayes, or non-parametric HMMS (17.5.5.1 \cite{murphy2012machine}).
\end{comment}
\begin{proof}
Consider, without loss of generality, a Viterbi graph where we examine two consecutive time-steps $(t-1, t)$ along two possible latent states $l,k$ (the analysis generalizes to HMMs with more states). We also assume $\forall i, i = \arg\max_{j}A_{ij}$; that is,  all hidden states states tend to self-transition. Also at time $t-1$:
\begin{align}
	\delta_{t-1}(l) > \delta_{t-1}(k)\label{delta_t_minus_1_relation}.
\end{align}
We also define the following symbol:
\begin{align}
	w_{ji}(t) = A_{ji}*b_{i}(y_t).
\end{align}
Due to our first assumption, we have: $\max_{j}w_{ji}(t) = w_{ii}(t)$. Then, at time $t$, the \(\delta\) values are:
\begin{align}
	\delta_{t}(l) &= max(\delta_{t-1}(l)*w_{ll}(t), \delta_{t-1}(k)*w_{kl}	(t))\label{delta_t_l}\\
	\delta_{t}(k) &= max(\delta_{t-1}(l)*w_{lk}(t), \delta_{t-1}(k)*w_{kk}(t))
\end{align}
According to \ref{delta_t_minus_1_relation} and our max weight formulation, the max function in \ref{delta_t_l} is: $\delta_{t}(l) = \delta_{t-1}(l)*w_{ll}(t)$. So, the max state $l$ at time $t-1$ will contribute to itself instead of $k$ at time $t$. Therefore, there is only one condition under which the Viterbi sequence breaks: 
\begin{align*}
	\delta_{t}(k)=\delta_{t-1}(k)*w_{kk}(t) \quad \textrm{and} \quad \delta_{t}(k)>\delta_{t}(l)
\end{align*}
In other words, given our original assumption, the Viterbi sequence breaks if the following inequalities are met:
\begin{align*}
	\delta_{t-1}(k)*w_{kk}(t)&>\delta_{t-1}(l)*w_{lk}(t) \\
	\quad &\textrm{and} \quad \\
	\delta_{t-1}(k)*w_{kk}(t)&>\delta_{t-1}(l)*w_{ll}(t)
\end{align*}
In ratio form:
\begin{align}
	\frac{w_{kk}(t)}{w_{lk}(t)}&> \frac{\delta_{t-1}(l)}{\delta_{t-1}(k)} 			\label{ratio_test_1}\\
	\quad &\textrm{and} \quad \nonumber \\
	\frac{w_{kk}(t)}{w_{ll}(t)}&>\frac{\delta_{t-1}(l)}{\delta_{t-1}(k)}	 		\label{ratio_test_2}
\end{align}

If an observation is emitted by state $l$ and it is not undergoing a state switch, ($\delta_{t-1}(l)>\delta_{t-1}(k)$ and $w_{ll}(t)>w_{kk}(t)$), inequality \ref{ratio_test_2} fails and the Viterbi sequence does not break.

When we transition from state $l$ to $k$ and begin emitting $w_{kk}(t)>w_{ll}(t)$. However, the momentum in $\delta_{t}(k)$ and $\delta_{t}(l)$ prevent their inequality relationship to switch. Nonetheless, after $p$ time steps, the inequality $\delta_{t-1}(l) > \delta_{t-1}(k)$ becomes $\delta_{t-1+p}(l) < \delta_{t-1+p}(k)$. The latter is reasonable when state $k$ has been emitting for some time. It is before this time $t-1+p$ that inequalities \ref{ratio_test_1} and \ref{ratio_test_2} are met. To see why, one will notice the left hand side inequalities in \ref{ratio_test_1} and \ref{ratio_test_2} are larger than 1 while the right hand side ratios become smaller than 1 at time $t-1+p$, thus their inequality relationships must've swapped before this time. Note that the order in which inequalities \ref{ratio_test_1} and \ref{ratio_test_2} are met matters. If \ref{ratio_test_2} is met but \ref{ratio_test_1} is not, a clean cut occurs since we have:
\begin{align}
	\delta_{t-1}(k)*w_{kk}(t)&<\delta_{t-1}(l)*w_{lk}(t)\label{clean_cut_ineq_1}\\
	\quad &\textrm{and} \quad\nonumber \\
  	\delta_{t-1}(k)*w_{kk}(t)&>\delta_{t-1}(l)*w_{ll}(t)\label{clean_cut_ineq_2}
\end{align}
Eqtn. \ref{clean_cut_ineq_2} asserts a switch from $l$ to $k$. Eqtn. \ref{clean_cut_ineq_1} states that $\delta_{t-1}(l)$ contributes to $\delta_{t}(k)$, implying the previous max state contributes to the next max state. And since the max state has changed, the roles of $l$ and $k$ swap and inequality \ref{ratio_test_2} begins to fail and renders sequence break unattainable. This yields a clean transition cut. If \ref{ratio_test_1} and \ref{ratio_test_2} are met simultaneously, a sequence break occurs. But since \ref{ratio_test_2} is met, the states switch and the roles of $l$ and $k$ swap and preclude a further sequence break.

If, let's say at time $t_a$, inequality \ref{ratio_test_1} is met but \ref{ratio_test_2} is not, a future sequence break is destined to happen at the future moment when \ref{ratio_test_2} is met, let's say at time $t_b$. When that sequence break occurs, the history between $t_a$ and $t_b$ flips and after the sequence breaks the state transitions from $l$ to $k$. Again, roles switch and no further sequence breaks occur. Above all, we can safely conclude that the during the execution of the stable period of a hidden state, no sequence break will occur. It is only during state transitions that a sequence could break and even if it does, it only last for a single time step--the time step when inequality \ref{ratio_test_2} is met. The analysis extends to HMM models with more than 2 hidden states, so long as we we apply the analysis to pairs of hidden states composed of the current max state and another non-max state.

Theorem \ref{Viterbi_path_presupposition}, is amply supported by our HMM models dynamics and evidenced in the color-coded Viterbi path plot of Fig. \ref{fig:Viterbi_triangle_emissionprob_gradient}a where Viterbi paths grow stably over time (rows). During state transitions, \textit{negligible sequence breaking occurs for one time-step} and quickly returns to the stable growth of Viterbi paths. Fig. \ref{fig:Viterbi_triangle_emissionprob_gradient} shows three well formed triangles indicating 3 skills with stable dynamics and clean transitions. Data and analysis supporting our finding is included in our supplemental document in \cite{2018ROMAN-Luo-EventDetection_supplementalURL}.
\end{proof}
\begin{corollary}
\label{loglik_gradient_depend_on_emissionprob_and_transmat_inference}
Given Theorem \ref{Viterbi_path_presupposition} the gradient of the log-likelihood of the forward algorithm (from now on referred to as the forward gradient) will depend solely on the latest emission probabilities and the transition matrix. 
\end{corollary}
\begin{comment}
This corollary is supported by our HMM model as evidenced in Fig. \ref{fig:Viterbi_triangle_emissionprob_gradient}b,c. In Fig. \ref{fig:Viterbi_triangle_emissionprob_gradient}b, the 3 colored curves represent the emission probabilities of corresponding hidden states. They are clearly distinct. Furthermore, the 3 curves' maxima match with the forward gradients as seen in Fig. \ref{fig:Viterbi_triangle_emissionprob_gradient}c. This relationship manifests that the forward gradient is directly related to the latest emission probabilities.
\end{comment}
\begin{proof}
According to the log-exp-sum trick \cite{murphy2012machine}, the approximation \(\log{\sum_{i=1}^{N}{\exp(y_i)}}\approx\max_{i\in\{1, \cdots, N\}}{y_i}\) is best approached for larger values of $y_i$. Applying this approximation to Eqtn. \ref{eq:sum_of_log_alpha} and Eqtn. \ref{eq:log_alpha}, which is supported by Theorem \ref{Viterbi_path_presupposition}, we have:
\begin{equation}
	L_t \approx \max_{i\in\{1, \cdots, N\}} ( {\log{{\alpha_i}(t)}} )
    \label{eq:approximated_L_t}
\end{equation}
\begin{equation}
	\begin{aligned}
    \log{ {\alpha_i}(t) } & \approx \log{ {b_i}(y_{t}) } +\\
    &\quad \max_{j \in \{1, \cdots, N\} } ( {\log{ {{\alpha_j}(t-1) } + \log{A_{ji}}}} )
    \label{eq:approximated_log_alpha_i}
    \end{aligned}
\end{equation}
Substitute Eqtn. \ref{eq:approximated_log_alpha_i} into Eqtn. \ref{eq:approximated_L_t}, rename \(i\) and \(j\) to \(z_t\) and \(z_{t-1}\), then recursively decompose \(\log\alpha\), we have:
	\begin{align}
    L_t&=\max_{{z_t}\in\{1, \cdots, N\}}({\log{{\alpha_{z_t}}(t)}}) \nonumber \\
  	&=\begin{aligned}[t]
    &\max_{{z_t}\in\{1, \cdots, N\}}(\log{{b_{z_t}}(y_{t})}+\\
    &\quad\max_{{z_{t-1}}\in\{1, \cdots, N\}}(\log{A_{{z_{t-1}}{z_t}}}+		{\log{{{\alpha_{z_{t-1}}}(t-1)}}}))
  	\end{aligned} \nonumber \\
	&=\begin{aligned}[t]
  	&\max_{{z_t}\in\{1, \cdots, N\}}(\log{{b_{z_t}}(y_{t})}+\\
  	&\quad\max_{{z_{t-1}}\in\{1, \cdots, N\}}(\log{A_{{z_{t-1}}{z_t}}}+\log{b_{z_{t-1}}(y_{t-1})}+\\
  	&\quad\quad\ddots\\
  	&\quad\quad\quad\max_{{z_{1}}\in\{1, \cdots, N\}}(\log{A_{{z_{1}}{z_{2}}}}+{\log{{\pi_{z_1}{b_{z_1}(y_1)}}}})\cdots))
  	\end{aligned}
    \label{eq:loglik_computation_recursion}
  	\end{align}
Eqtn. \ref{eq:loglik_computation_recursion} is the log version of Eqtn. \ref{eq:Viterbi_recursion}. This suggests that, after approximations by equations \ref{eq:approximated_L_t} and \ref{eq:approximated_log_alpha_i}, the computation of the log-likelihood is the same as the computation of the Viterbi path using the Viterbi algorithm.

Now, since Theorem \ref{Viterbi_path_presupposition} shows that in general the Viterbi path at time \(t\), \(\hat{z}_{1:t}\), expands on the Viterbi path at time \(t-1\), \(\hat{z}_{1:t-1}\), we have:
\begin{align}
	L_t&=\begin{aligned}[t]
  	&\max_{{z_t}\in\{1, \cdots, N\}}(\log{{b_{z_t}}(y_{t})}+\\
  	&\quad\max_{{z_{t-1}}\in\{1, \cdots, N\}}(\log{A_{{z_{t-1}}{z_t}}}+{\log{{{\alpha_{z_{t-1}}}(t-1)}}}))
  	\end{aligned} \nonumber \\
  	&=\log{{b_{\hat{z}_t}}(y_{t})}+\log{A_{{\hat{z}_{t-1}}{\hat{z}_t}}}+L_{t-1}\label{eq:how_L_t_relates_to_L_t-1}. 
\end{align}
Then, the forward gradient can be derived from Eqtn. \ref{eq:how_L_t_relates_to_L_t-1} as:
\begin{align}
	\nabla L_t&=L_t-L_{t-1} =\log{{b_{\hat{z}_t}}(y_{t})}+\log{A_{{\hat{z}_{t-1}}{\hat{z}_t}}}
    \label{eq:gradient_of_L}. 
\end{align}
Eqtn. \ref{eq:gradient_of_L} supports Corollary   \ref{loglik_gradient_depend_on_emissionprob_and_transmat_inference}, where the forward gradient depends on the latest emission probability \({b_{\hat{z}_t}}(y_{t})\) and transition probability from hidden state \(\hat{z}_{t-1}\) to \({\hat{z}_t}\). Also, given that good HMM models have strong inertia (high probabilities of self-transitions), state-switching should be sparse and then \(\hat{z}_{t-1}\) will equal to \({\hat{z}_t}\) most of the time. 
\end{proof}
\begin{figure}[t]    
    \centering        
        \includegraphics[width=1\linewidth]{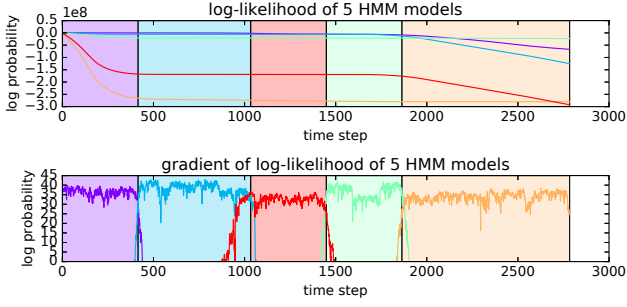}
        \caption{Identification ability of the HMM log-likelihood curve and the forward gradient. We consider 5 HMM models (color-coded) to represent one skill using 1 trial time series data. Background colors indicate true skill execution periods. The HMM log-likelihood curve does not correlate to skill execution periods, whilst the forward gradient curve shows strong correlations to skill execution periods.}
        \label{fig:test_if_gradient_can_detect_state_switch}        
\end{figure}
\section{Event Detection based on the Forward Gradient}
\subsection{Detect Normal Events: Skill Identification}
Corollary \ref{loglik_gradient_depend_on_emissionprob_and_transmat_inference} led us to design a new method for skill identification. If we use $n$ HMM models to represent $n$ robot skills, with observations coming from a certain skill, the HMM model corresponding to that skill $\hat{m}$, should output a value-increasing forward log-likelihood curve that is greater than the rest of the HMM models. This also means model $\hat{m}$ will output a larger forward gradient value compared to other models.

The forward gradient depends on the latest emission probabilities, which in turn depend on the latest observation. The largest probabilities and thus gradients will belong to the HMM model of a currently executing robot skill.
The key insight however is the inverse relationship: the use of the forward gradients to infer the currently executing skill. Fig. \ref{fig:test_if_gradient_can_detect_state_switch} validates the strong correlation between the forward gradient and skill observations. This is contrasted with the log-likelihoods of the observations $log \mbox{ } P(Y_{1:t} | \Pi)$ do not. 
The forward-gradient measure for skill identification is defined as follows: given $p$ skills $s_1:s_p$, we have HMM models $m_s$ for $s \in \{1, \cdots, p\}$  and an input time series $Y$, then the most probable skill $\hat{s}$ generating $Y[t]$ is inferred as:
\begin{align}
	\hat{s}=\arg\max_{s \in \{ 1, \cdots, n \}} ( \nabla L_t^{m_s}(Y) )
    \label{eqtn:max_forward_gradient}
\end{align}
where, $\nabla L_t^{m_s}(Y)$ is the forward gradient output by model $m_s$ at time $t$ computed using time series $Y$.
\begin{figure}[b]
	\centering
		\includegraphics[width=\linewidth]{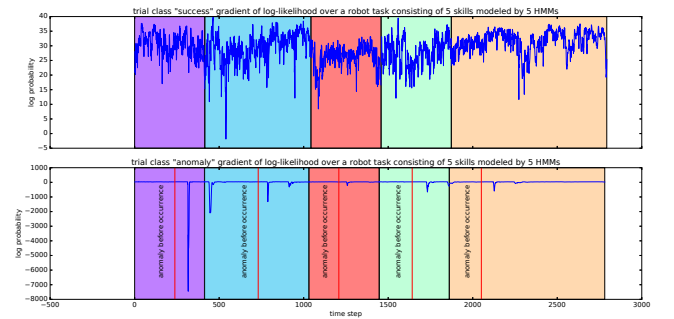}
		\caption{The log-likelihood gradient can be used  for anomaly identification. Top plot: nominal task with a steadily increasing log-likelihood that yields a positive gradient (ours ranges from 10-45 units in this trial). Bottom plot: trial with one anomaly per skill execution. Anomalies occur shortly after red vertical lines. Notice gradient drops after anomaly occurrence (range from -100's to -1000's).}
	\label{fig:gradient_comparison_between_success_and_anomaly_trials}        
\end{figure}
\subsection{Anomaly Detection}
We now build on the premise established in Eqtn. \ref{eqtn:max_forward_gradient}. Furthermore, consider a set of nominal observations for an executing skill, we know that the corresponding skill HMM model will output a value-increasing forward log-likelihood curve, and hence, a stable positive forward gradient. So, when an anomaly occurs, the forward gradient decreases significantly as illustrated in Fig. \ref{fig:gradient_comparison_between_success_and_anomaly_trials}. Given that anomalies influence the forward gradient value, we propose a gradient-based metric for HMM anomaly detection.

Consider an HMM model $m$ representing a skill $s$ with $n$ time-series trials $Y_i$ for $i \in \{1, \cdots, n\}$ collected from successful executions of skills $s \in S$. To detect anomalies in a new time series $Y$ we can first derive:

\[\nabla L_{max} = \max_{i\in\{1, \cdots, n\}}(\max_{t\in\{1, \cdots, T_i\}}(\nabla L^{m}_{t}(Y_i))),\]

\[\nabla L_{min} = \min_{i\in\{1, \cdots, n\}}(\min_{t\in\{1, \cdots, T_i\}}(\nabla L^{m}_{t}(Y_i))),\]

\[\nabla L_{range} = \nabla L_{max}-\nabla L_{min},\]

where $T_i$ is the time length of trial $Y_i$ and $\nabla L^{m}_{t}(Y_i)$ is the forward gradient output by model $m$ at time $t$ computed using time series $Y_i$. Then, we use an empirically-derived test to trigger an anomaly for $Y$:
\begin{align}
\nabla L^{m}_{t}(Y)<\nabla L_{min}-\frac{\nabla L_{range}}{2}.
\end{align}
This test detects if the gradient is an outlier compared with gradients of successful skill executions.
\begin{figure}[t]    
    \centering        
        \includegraphics[width=\linewidth]{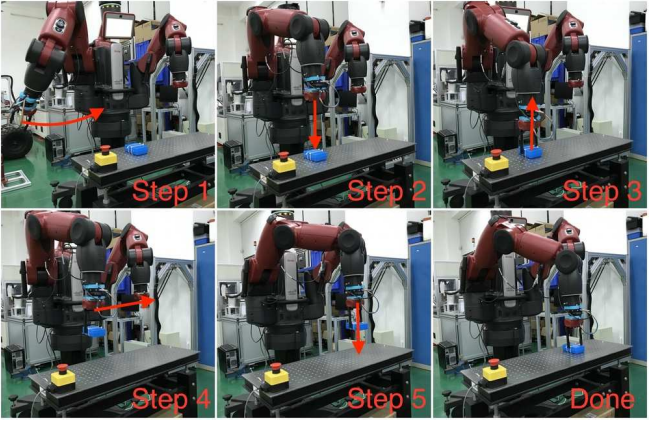}
        \caption{The Baxter humanoid robot performing a pick-and-place task. 5 independent skills used to perform the task. Executed skill motions are sketched with red arrows.}
        \label{fig:experiment_images}   
\end{figure}
\section{Experiments and Results} \label{sec:experiments}
As for experimental setup, a dual armed humanoid Baxter robot was used to perform a pick-and-place operation. The robot consisted of a Robotiq force-torque sensor and standard Baxter fingers. 5 nominal trials were used for training the HMM  model. In testing, 5 trials were used for skill identification and anomaly detection respectively. The pick-and-place task consists of 5 skills: 
\begin{enumerate*}[label=(\roman*)]
  \item hover over the picking position, 
  \item grasp the object,
  \item lift the object,
  \item hover to the placing position, and 
  \item place the object.
\end{enumerate*}
Fig. \ref{fig:experiment_images}, shows the experimental setup and the execution of the five skills. 

For training, the observation vector concatenates a 7-dimensional Cartesian end-effector pose and a 6-dimensional wrench. For each skill, we train corresponding HMM models using the Baum-Welch algorithm. The number of hidden states is selected such that emission probabilities are maximized leading to distinct and uniquely grouped hidden states.

For results reporting, we use the three factors identified by Pettersson's survey on Event Detection \cite{2005BotAut-Petterson-ExecutionMonitoringInRoboticsSurvey}. Namely: classification accuracy, robustness (false-positive rate), and reaction time  (the time it takes to identify a skill from the beginning of a skill execution). Note that for anomaly identification, internal and external perturbations are used including: unexpected movement of target object, object absence, slippery picks, and human collisions. 
\subsection{Skill Identification Performance}
Table \ref{tbl:confusion_matrix} presents The skill identification confusion matrix. Skills 1 and 3 were recognized with 100\% accuracy, 2 and 4 with 99\% accuracy, and skill 5 with the largest surface contacts with 94\% accuracy. Overall accuracy was 98.4\%. 
\begin{table}[t]
	\centering
	\caption{Skill identification confusion matrix for pick-and-place}
	\label{tbl:confusion_matrix}

\begin{tabular}{l|ccccc}
            & Skill 1 & Skill 2 & Skill 3 & Skill 4 & Skill 5 \\
\hline
Skill 1     & \red{1.00}    & 0.00    		& 0.00    		& 0.00    		& 0.00    \\
Skill 2     & 0.00    		& \red{0.99}   	& 0.00    		& 0.00    		& 0.01    \\
Skill 3     & 0.00    		& 0.00    		& \red{1.00}   	& 0.00    		& 0.00    \\
Skill 4     & 0.00    		& 0.00    		& 0.01    		& \red{0.99}   	& 0.00    \\
Skill 5     & 0.00    		& 0.04    		& 0.02    		& 0.00    		& \red{0.94}    \\
\end{tabular}
\end{table}
\subsection{Reaction Time Performance}\label{subsec:reaction_time}
\begin{figure}[h]    
    \centering        
        \includegraphics[width=\linewidth]{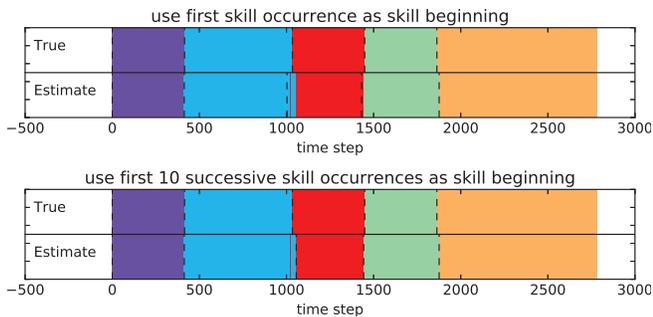}
        \caption{Two ways to determine the beginning of a skill while performing skill identification: (i) use the first skill occurrence, or (ii) use the first 10 successive occurrences. For each plot, the upper half shows the true skill at time $t$ and the lower half shows the prediction. Vertical dotted black lines mark the beginning of a skill. Predicted beginnings for the red skill vary significantly between the two criteria: the first plot determines that red skill begins as soon as one red skill estimation occurs even though that estimation is not stable.}
        \label{fig:skill_id_of_one_trial_compared_with_ground_truth}   
\end{figure}

In terms of reaction time, a percentage is computed to assess the time it takes for the identification to execute from the beginning of a skill. The reaction percentage, using t as ``true'', is computed as: \textit{offset = predicted-t\_beginning, length=t\_end-t\_beginning}, and \textit{reaction=offset/length}.
The closer the reaction percentage is to 0\% the better the identification method. A negative reaction percentage means the predicted start occurs earlier than ground truth, while a positive percentage implies a delayed identification. We assess two forms to determine the beginning of a skill as illustrated in Fig. \ref{fig:skill_id_of_one_trial_compared_with_ground_truth}: (i) use the ``first skill'' occurrence, or (ii) use the ``fist 10 successive skill'' occurrences. The reaction percentage for these two formats is found in Table \ref{tbl:reaction_percentage}.The average reaction time for absolute values (i.e. looking at the average time difference of the prediction, whether early or late) the ``first skill'' is 2.70\% across all skills and the average reaction time for the ``first 10 skills'' is 0.97\%. Between the two measures, we have a total average of 1.84\%. 
\begin{table}[h]
\centering
    \caption{Reaction Time as a Duration Percentage of a Skill.}
    \label{tbl:reaction_percentage}
    \begin{tabular}{cc|c|c|c|clllll}
    \multirow{2}{*}{Method}                                                                                         & \multicolumn{5}{c}{Average Reaction Percentage} &  &  &  &  &  		\\ 
    \cline{2-6}                                                                                                     & skill 1  & skill 2 & skill 3 & skill 4 & skill 5 &  &  &  &  &  		\\ 
    \cline{1-6}
    \begin{tabular}[c]{@{}c@{}}first skill occurrence
    \end{tabular}                   
    & 0.00\%   & 0.23\%  & -3.23\% & \red{0.28\%}  & -9.77\% &  &  &  &  &  \\ 			
    \begin{tabular}[c]{@{}c@{}}first 10 occurrences 
    \end{tabular} 
    & 0.00\%   & 0.23\%  & \red{-2.36\%} & 0.65\%  & \red{1.61\%}  &  &  &  &  &  
    \end{tabular}
\end{table}
\subsection{Anomaly Detection Performance}
For anomaly detection performance of our gradient-based method, we use two environments: (i) anomaly identification as it occurs and any false-positives before an actual anomaly occurs, and (ii) an external collision is given to the robot to trigger a recovery. Then, when the robot completes its recovery behavior, we count how many false-positives are triggered before moving to the next skill execution. The robot recovery behavior is detailed in \cite{2018ROMAN-Wu-RecovExtDist_StateDep}. Five nominal and five anomalous trials are used for the analysis. The results are compared with two other baseline methods: the magnitude-based metric from Sec. \ref{subsec:previ_skill_id}, and the derivative-of-difference metric from Sec. \ref{subsec:PrevAnomalyIdentMethods}. 

The five anomalous trials contain a total of 14 anomalies, consisting of: 
\begin{enumerate*}[label=(\roman*)]
\item one anomaly caused by the displacement of the target object
\item one anomaly caused by no target object
\item two anomalies caused by slippery picks
\item five anomalies caused by human collisions to the robot gripper during each skill execution
\item five anomalies caused by human collisions to the robot arm during each skill execution
\end{enumerate*}.
\subsubsection{Pre-Recovery Performance}\label{subsubsec:pre_rec_perf}
\begin{figure}[hb]
	\centering
    \includegraphics[width=1\linewidth]{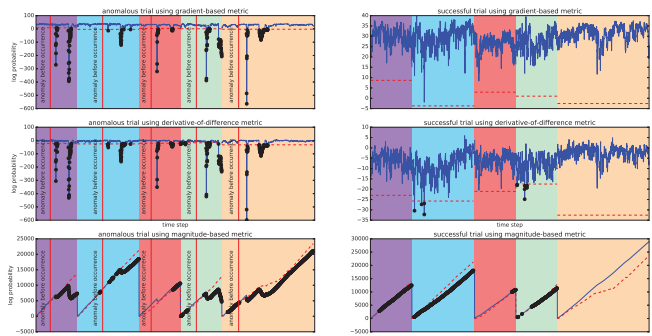}
    \caption{Two trials: nominal on the left and anomalous on the right. Detection metrics: gradient-based, derivative-of-difference, and magnitude-based arranged by row. Blue curves show metric values, red dashed lines thresholds, and black circled markers triggered anomalies. Markers occur from the beginning to the end of a skill. Vertical red line are false positives (no anomaly). Markers to the right of the red vertical line are true positives and render markers remaining within the skill as trivial. If no marker is shown to the right of the vertical line, it is treated as a false negative. For successful trials, closely located markers are grouped as a false positive.}
    \label{fig:metrics_comparison_img}
\end{figure}

For pre-recovery performance computation, during each trial, we record the anomalies triggered by the testing metric and count its true positives, false positives and false negatives as illustrated in Fig. \ref{fig:metrics_comparison_img}. 
Our result summary is shown in Table \ref{metrics_performance}. The result shows our proposed forward gradient detected all anomalies and triggered no false positives or false negatives. The other two baseline methods suffer from false positives though they deliver high true positives.
\begin{table}[t]
	\centering
	\caption{Anomaly Detection Metrics Performance Comparison}
	\label{metrics_performance}
\begin{tabular}{c|c|c|c}
	\multirow{2}{*}{detection metric}  &  \multicolumn{3}{c}{Micro}   \\ 
\cline{2-4} 
		& F-score & Precision & Recall  \\ 
\hline
\begin{tabular}[c]{@{}c@{}}Proposed forward-gradient\\ measure\end{tabular}  & 100\%   & 100\%    & 100\%   \\ 
\hline
\begin{tabular}[c]{@{}c@{}}Derivative-of-difference\\ measure\end{tabular} & 82.35\% & 70\%     & 100\%   \\ 
\hline
\begin{tabular}[c]{@{}c@{}}Magnitude-based\\ measure\end{tabular}          & 60.47\% & 44.83\%  & 92.86\%
\end{tabular}
\end{table}
\subsubsection{Post-Recovery Performance}\label{subsubsec:post_rec_perf}
For post-recovery performance metrics, we trigger an intentional anomaly, after recovery is completed, we count any false-positives before next skill execution. Both the forward gradient method and the derivative-of-difference method had not false-positives. Whilst the magnitude-based metric had more than 10 and prevented the system from continuing its task execution.


\subsubsection{Comparison with Related Works}
Comparisons across works is challenging as results use different formats across experiments. Table \ref{tbl:performance_comparison} is an effort to harmonize results across related papers. The comparison should be done loosely as different tasks (small levels of contact vs. large levels of contacts, structured environment vs. unstructured environment) present different challenges to event detection. For skill identification, our current approach ranks 2nd behind the tool breakage work that identified anomalies in structured milling processes. Our work did better than \cite{2017iros-rojas-onlinewrenchintrospection} and \cite{2013IROS-DiLello-BayesianContFaultDetection}, albeit these works modeled more complex dynamical phenomena. Similar statements can be made about anomaly identification. As for reaction times, our approach offers about double the speed-up compared to the only other work that reported this number. In conclusion, based on internal and external evidence, we hold that our measure is the most robust, stable, and fastest measure reported to date.
\begin{table}[tb]
\centering
\caption{Skill and anomaly identification, and reaction time comparison for state-of-the-art event-detection methods.}
\label{tbl:performance_comparison}
\begin{tabular}{lll}
technique	                                                               & ID Accuracy    											 &  \\
\hline
AFF/DCC/CSM/SVM				\cite{2013MICCAI-Ahmidi-StringMotifDescrToolMotion_SkillGestures}				& 84.66\%
\footnote{Average of Table 3, k-fold: Gesture classification performance (novice, inter, and expert and both macro/micro levels).}    	 &  \\
sHDP-HMM 					\cite{2013IROS-DiLello-BayesianContFaultDetection}                              & 89.50\%                    &  \\
RCBHT w/ multiclass SVM 	\cite{2017iros-rojas-onlinewrenchintrospection}                                 & 97.00\%                    &  \\
HMM w/GradientBased Measure {[}current{]}                                   								& 98.40\%             		 &  \\
Tool breakage SVM 			\cite{2005IJMTM-Cho-ToolBreakageDetection}                                      & \red{99.38\%}              &  \\
\\
technique                                                                   & anomalyID Accuracy 										 &  \\
\hline
HMM,varying threshold 		\cite{2016ICRA-Park-MultiModalMonitoringAnomalyDet_RobotManip}                 	& $\sim$ 80.00\%             &  \\
MLP 						\cite{2017IROS-Park-MultiModalAnomalyClassifFeeding}							& 83.27\%                    &  \\
sHDP-VAR-HMM,mag metric 	\cite{2017humanoids-rojas-shdp-var-hmm}		                   					& $\sim$ 85.00\%             &  \\
sHDP-HMM 					\cite{2013IROS-DiLello-BayesianContFaultDetection}                              & 87.50\%                    &  \\
RCBHT w/ multiclass SVM 	\cite{2017iros-rojas-onlinewrenchintrospection}              	                & 97.00\%                    &  \\
HMM, gradient metric (current)		           	                        									& \red{100.00\%}             &  \\
\\
technique	                                                               & reaction time            			                         &  \\
\hline
sHDP-VAR-HMM,mag metric 	\cite{2017humanoids-rojas-shdp-var-hmm}	              							& 3.70\%
\footnote{Average reaction time for their best multimodal setup.}																		&  \\
HMM, gradient metric (current)             										                      		& \red{1.84\%}                & 
\end{tabular}
\end{table}
\section{Discussion} \label{sec:discussion}
This work presented a theoretically derived event detection measure useful for nominal and anomalous behavior identification, even in post-recovery actions. Our results showed very strong performance compared with state-of-the-art results across the board.
More experimental validation is certainly necessary: both in number of trials and robotic tasks. This work also remains to be tested in the area of anomaly classification. The latter is concerned not only with the identification problem with the grouping of anomaly types which is more challenging. We anticipate working in conjunction with machine or deep learning models for the classification of this signals. Some works \cite{2013IROS-DiLello-BayesianContFaultDetection,2014Humanoids-Rojas-ContextualizedEArlyFailureCharac,2017IROS-Park-MultiModalAnomalyClassifFeeding} already provide some characterizations.
\section{Conclusion} \label{sec:conclusion}
We presented an accurate, robust, fast, and versatile measure for skill and anomaly identification. The gradient-based measure devised through a theoretical proof established the link between the derivative of the HMM logarithm of the filtered belief state and the latest emission probabilities. We established that the latest emissions directly affect the gradient and that the key insight was the inverse relationship, which enabled nominal and anomalous identification with strong guarantees. The measure had strong performance for skill and anomaly identification including in post-anomaly-recovery scenarios. The measure proved to be both versatile and fast-acting and broadly applicable to event-detection if using HMM-based methods. 
\section{Acknowledgements} \label{sec:Acknowledgements}
This work is supported by ``Major Project of the Guangdong Province Department for Science and Technology (2014B090919002), (2016B0911006) and by the National Science Foundation of China (61750110521).''
\bibliographystyle{IEEEtran}
\bibliography{IEEEabrv,Xbib}

\end{document}